\documentclass[journal]{IEEEtran}
\usepackage{amsfonts}
\usepackage{amsmath}
\usepackage[numbers,sort&compress]{natbib}
\usepackage{url}
\usepackage{bm}
\usepackage{graphicx}
\usepackage[dvipsnames]{xcolor}
\usepackage{booktabs}
\usepackage{subfigure}

\ifCLASSINFOpdf
\else
\fi

\begin{document}

\title{IM2HEIGHT: Height Estimation from Single Monocular Imagery via Fully Residual Convolutional-Deconvolutional Network}
\author{Lichao~Mou,~\IEEEmembership{Student Member,~IEEE} and
       ~Xiao~Xiang~Zhu,~\IEEEmembership{Senior Member,~IEEE}
\thanks{Manuscript received September 18, 2017.}
\thanks{This work is jointly supported by the China Scholarship Council, the European Research Council (ERC) under the European Union's Horizon 2020 research and innovation programme (grant agreement No [ERC-2016-StG-714087], Acronym: \textit{So2Sat}), and Helmholtz Association under the framework of the Young Investigators Group ``SiPEO'' (VH-NG-1018, www.sipeo.bgu.tum.de).}
\thanks{The authors are with the Remote Sensing Technology Institute (IMF), German Aerospace Center (DLR), Germany and with Signal Processing in Earth Observation (SiPEO), Technical University of Munich (TUM), Germany (e-mails: lichao.mou@dlr.de; xiao.zhu@dlr.de).}
       }% <-this % stops a space

\markboth{Submitted to IEEE Transactions on Geoscience and Remote Sensing on September 18, 2017}%
{Shell \MakeLowercase{\textit{Mou et al.}}: IM2HEIGHT}

\maketitle

\begin{abstract}
  In this paper we tackle a very novel problem, namely height estimation from a single monocular remote sensing image, which is inherently ambiguous, and a technically ill-posed problem, with a large source of uncertainty coming from the overall scale. We propose a fully convolutional-deconvolutional network architecture being trained end-to-end, encompassing residual learning, to model the ambiguous mapping between monocular remote sensing images and height maps. Specifically, it is composed of two parts, i.e., convolutional sub-network and deconvolutional sub-network. The former corresponds to feature extractor that transforms the input remote sensing image to high-level multidimensional feature representation, whereas the latter plays the role of a height generator that produces height map from the feature extracted from the convolutional sub-network. Moreover, to preserve fine edge details of estimated height maps, we introduce a skip connection to the network, which is able to shuttle low-level visual information, e.g., object boundaries and edges, directly across the network. To demonstrate the usefulness of single-view height prediction, we show a practical example of instance segmentation of buildings using estimated height map. This paper, for the first time in the remote sensing community, attempts to estimate height from monocular vision. The proposed network is validated using a large-scale high resolution aerial image data set covered an area of Berlin. Both visual and quantitative analysis of the experimental results demonstrate the effectiveness of our approach.
\end{abstract}


\IEEEpeerreviewmaketitle

\section{Introduction}
\label{sec:intro}
Estimating height is of great importance in understanding geometric relations within a scene. Many challenging remote sensing problems, on the other hand, have proven to benefit from the rich representations of objects and their environment provided by the height information, to name a few, semantic labeling~\cite{Volpi17,Audebert17,DFC15,Marmanis16} and change detection~\cite{Qin15,Qin16}.
\par
The highly developed laser sensors like LiDAR nowadays have made the acquisition of DSM data affordable, but often such height information is not always available, especially working on a huge number of historical remote sensing images. In addition, while there is much prior work in remote sensing field on estimating height based on the idea of stereo matching, which leverages camera motion to estimate camera poses through different temporal interval and, in turn, estimate height via triangulation from pairs of consecutive views, there has been fairly little on estimating height from a \emph{single} remote sensing image. Whereas the monocular case often arises in practice.
\par
Unfortunately, height estimation from monocular vision is a technically ill-posed problem, as one captured remote sensing image may correspond an infinite number of possible real world scenarios, i.e., there is an inherent ambiguity in mapping an intensity or color measurement into a height value. Yet it is not difficult for humans to infer the underlying 3D structure from a single remote sensing image, which remains a challenging task to develop a vision algorithm to do so by exploiting monocular cues alone. In this work, a correct height map means one is physically plausible in real world.

\subsection{Related Work}
\emph{\textbf{Depth Estimation in Computer Vision.}} Directly related to our work are several methods of estimating depth map from a single image in computer vision field. Traditionally, this problem was solved by works making use of hand-crafted visual features and probabilistic graphical models (PGMs). And most approaches rely on strong assumptions about scene geometry. Saxena \emph{et al.}~\cite{Saxena05} use a discriminatively-trained Markov Random Field (MRF) that infers depth from a single monocular image by incorporating local and global image features, in which superpixels are introduced to enforce neighboring constraints. Their work has been later extended to the Make3D~\cite{Saxena08} system for 3D scene structure generation. \cite{Hoiem05}, where authors make an attempt at classifying image regions into geometric structures (e.g., sky, vertical, and ground) instead of explicitly predicting depth. By doing so, a simple 3D model of the scene can be obtained. Liu \emph{et al.}~\cite{Liu10} do not map from hand-crafted features to depth directly, but instead first perform a semantic segmentation of the scene and use the semantic labels to guide the depth prediction. In this respect, depth can be more readily estimated by measuring the difference in appearance with respect to a given semantic class. In~\cite{Ladicky14}, authors show how to simultaneously integrate semantic segmentation with single-view depth estimation to improve performance.
\par
Another view on depth estimation can be found when considering transfer strategy: a feature-based matching between a query image and a pool of images for which the depth is known is first performed to search the nearest neighbors, and the retrieved depth counterparts are then warped and combined to produce the final depth map for the query image. Konrad \emph{et al.}~\cite{Konrad12} make use of a kNN transfer mechanism based on histograms of oriented gradient (HOG) feature to estimate depth by computing a median over the retrieved candidate depth maps. In~\cite{Liu14}, authors consider monocular depth estimation a discrete-continuous optimization problem, where the continuous variables encode the depth of the superpixels in the query image, and the discrete ones represent relationships between neighboring superpixels. Thus they can formulate this task as inference in a high-order Conditional Random Field (CRF). These transfer mechanism-based approaches, however, require the entire datasets to be available at runtime and perform expensive alignment process.
\par
More recently, as an important branch of the deep learning family, convolution neural networks (CNNs), have been fast emerging as the leading machine learning methodology and become the model of choice in many fields. For instance, CNNs have been proven to be good at extracting mid- and high-level abstract feature representations from raw images for classification purpose, by interleaving convolutional and pooling layers, i.e., spatially shrinking the feature maps layer by layer, and recently proposed network architectures also allow for dense per-pixel predictions like semantic segmentation~\cite{FCN,Noh15} and super-resolution~\cite{Dong16,Shi16}. A first attempt in exploiting CNN for depth estimation can be found in the work of Eigen \emph{et al.}~\cite{Eigen14}, where authors achieve this through the use of two CNNs, one that regresses a global depth structure from a single image, and another that refines it locally at finer resolution. In~\cite{Eigen15}, that work is later extended by including the idea of predicting depth jointly with the semantic labeling and surface normal estimation by a multi-scale CNN architecture, which helps obtaining more fine-grained depth maps. Moreover, another line of works studies ways of combining CNNs and graphical models like CRF for single-view depth estimation, where CNNs are usually used to extract relevant features. For example, Liu \emph{et al.}~\cite{LiuPAMI16} employ a CRF to model the relations of neighboring superpixels and learn the potentials (both unary and binary) in a unified CNN framework.
\par

\begin{figure}
\begin{minipage}[t]{\columnwidth}
\centering
\subfigure[]{
\includegraphics[width=0.48\columnwidth]{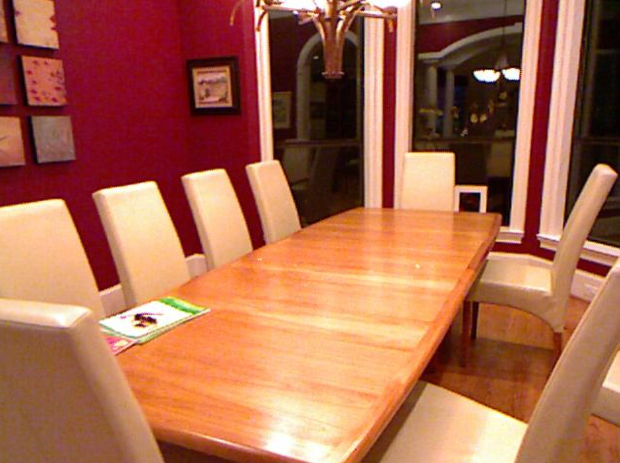}}
\subfigure[]{
\includegraphics[width=0.48\columnwidth]{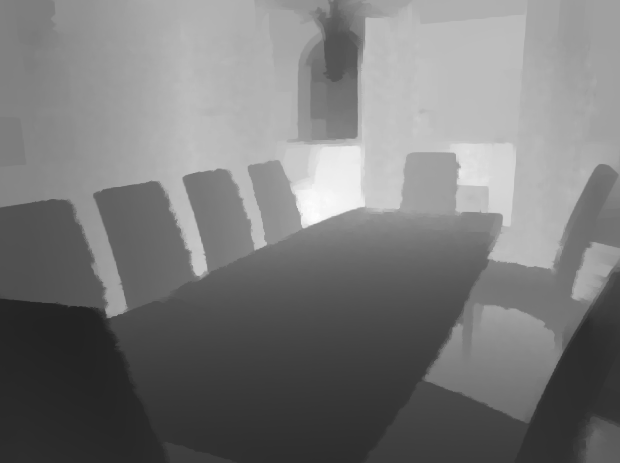}}
\end{minipage}
\begin{minipage}[t]{\columnwidth}
\centering
\subfigure[]{
\includegraphics[width=0.48\columnwidth]{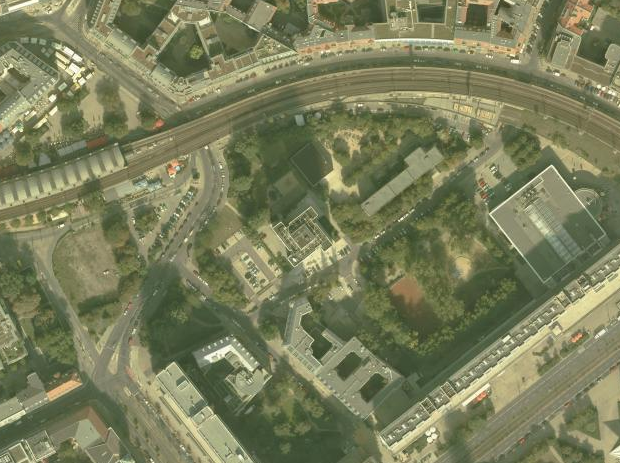}}
\subfigure[]{
\includegraphics[width=0.48\columnwidth]{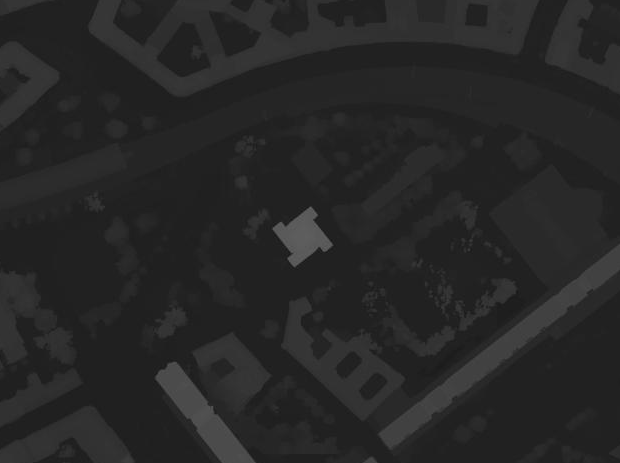}}
\end{minipage}
\renewcommand{\figurename}{Fig}
\caption{\label{fig:example} Examples of depth estimation data in computer vision and height prediction data in remote sensing. \textbf{Top:} (a) an RGB image from NYU-Depth V2 data set~\cite{Silberman12} and (b) its corresponding depth map gathered by a Kinect. \textbf{Bottom:} (c) a remote sensing image and (b) its DSM data used in this work.}
\end{figure}

\emph{\textbf{Height Prediction in Remote Sensing.}} For depth/height estimation from single monocular images, the useful cue to be taken into account is the \emph{contextual information} that includes occlusion, interposotion, texture gradient, and texture variations~\cite{SaxenaIJCV08,Bulthoff98}. In comparison with images used for depth estimation in computer vision field, remote sensing images have several unique characteristics which bring in new challenges for height prediction:
\begin{itemize}
  \item Remotely sensed images are often orthographic, which leads to the fact that extremely limited \emph{contextual information} is available.
  \item In remote sensing, limited spatial resolution, relatively large area coverage, and a number of tiny objects represent often a problem for height prediction task.
\end{itemize}
Fig.~\ref{fig:example} shows the differences between the depth estimation data and the height prediction data.
\par
Unlike single-view depth estimation in computer vision, height prediction from a single monocular image has rarely been addressed in the remote sensing community so far. In a pioneer work, Srivastava \emph{et al.}~\cite{Srivastava17} use a joint loss function in a CNN, which is a linear combination of a semantic labeling loss and a regression loss minimizing height prediction errors. The network can be trained by traditional back-propagation by alternating over the two losses. However, note that in the training phase this model needs pixel-wise labeled segmentation masks as input, while obtaining massive amount of manually-labeled masks is very expensive and time consuming. In contrast, along with the development of sensor technology, DSM data is now widely accessible at a reasonable cost.

\begin{figure*}[!t]
\centering
\includegraphics[width=\linewidth]{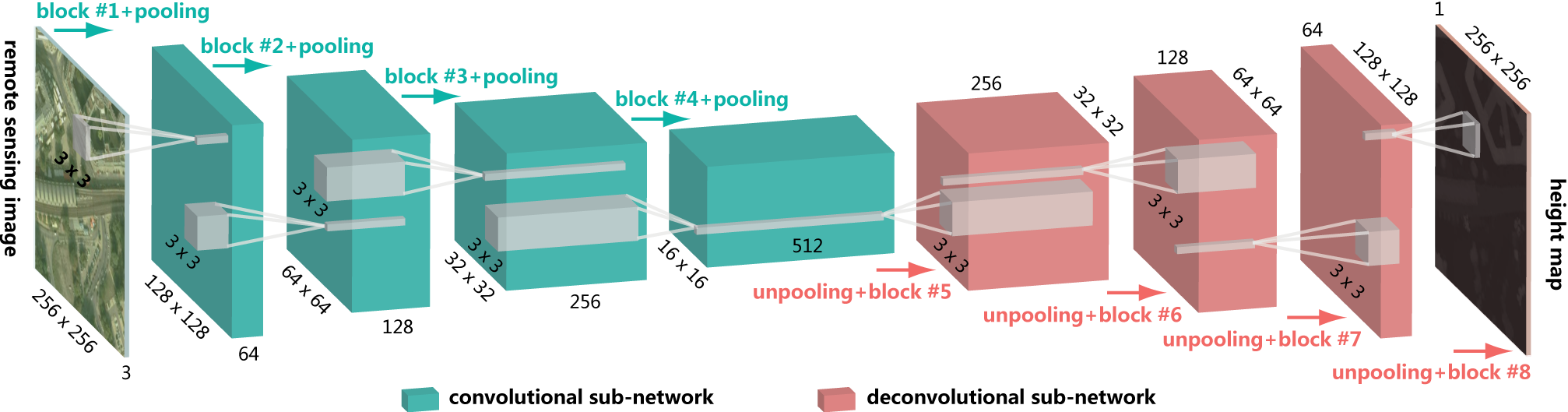}
\renewcommand{\figurename}{Fig}
\caption{\label{fig:net} Overview of the convolutional-deconvolutional architecture. The network is composed of two parts, i.e., convolutional sub-network and deconvolutional sub-network. The former corresponds to feature extractor that transforms the input remote sensing image to high-level multidimensional feature representation, whereas the latter plays the role of a height generator that produces height map from the feature extracted from the convolutional sub-network; see Fig.~\ref{fig:block} for details on block.}
\end{figure*}

\subsection{Contributions}
In this paper we propose to learn the mapping between a single monocular remote sensing image and its corresponding height map, by exploiting an end-to-end network, and unlike~\cite{Srivastava17}, only DSM is used as ground truth to train the network. More specifically, we directly regress on the height making use of a network with two components: one that first transforms the input image into a condensed, abstract high-level feature representation, then a second that estimates the height map of the scene from this encoded feature. In detail, our work contributes to the literature in three major aspects:
\begin{itemize}
  \item In this paper we address a very novel problem in the remote sensing community, namely height estimation from a single monocular remote sensing image. Unlike~\cite{Srivastava17}, we only make use of DSM data as ground truth for training, and no additional data like pixel-wise labeled mask is used.
  \item We propose an end-to-end deep residual network, which is composed of a convolutional sub-network and a deconvolutional sub-network, as well as a skip connection. Learning such a residual convolutional-deconvolutional network architecture for pixel-wise prediction remote sensing tasks has not been well investigated yet to the best of our knowledge.
  \item To further assess the usefulness of the single-view height prediction, we show an application, instance segmentation of buildings from the predicted height map. Most instance segmentation approaches usually rely on strong supervision for training, i.e., pixel-level segmentation masks, while labeling a considerable number of pixel-level masks is expensive. In this application, we give a different perspective to achieve this task.
\end{itemize}
\par
The paper is organized as follows. After the introductory Section~\ref{sec:intro} detailing depth/height estimation from a single monocular image, we enter Section~\ref{sec:method} dedicated to the details of the proposed fully residual convolutional-deconvolutional network for height estimation. Section~\ref{sec:exp} then provides the network setup, experimental results, and an application. Finally, Section~\ref{sec:con} concludes the paper.

\section{Methodology}
\label{sec:method}
Denote by $(\bm{x},\bm{y})$ random variables representing a remote sensing image and its corresponding DSM data, and denote their joint probability distribution by $p(\bm{x},\bm{y})=p(\bm{x})p(\bm{y}|\bm{x})$. Here $p(\bm{x})$ is the distribution of remote sensing images and $p(\bm{y}|\bm{x})$ is the distribution of DSM maps given remote sensing images. Ideally our aim is to find $p(\bm{y}|\bm{x})$, but direct application of Bayes' theorem is not feasible. Fortunately, as a special case, $\bm{y}$ may be a deterministic function of $\bm{x}$. Therefore in this paper we resort to a point estimate a mapping $f:\bm{x}\rightarrow \bm{y}$, which minimizes the following objective function:
\begin{equation}\label{eq:1}
\mathbb{E}_{\bm{x},\bm{y}}l(\bm{y}-f(\bm{x}))\,,
\end{equation}
where $l(\cdot)$ is a loss function.
\par
The minimizer of this loss is the conditional expectation:
\begin{equation}
\hat{f}(\bm{x}_0)=\mathbb{E}_{\bm{y}}[\bm{y}|\bm{x}=\bm{x}_0]\,,
\end{equation}
that is the expected height map.
\par
Given the training set of remote sensing images and their DSM data $\{\bm{x}_i,\bm{y}_i\}$, we learn the weights $\bm{\Theta}$ of $f(\bm{x},\bm{y})$ to minimize a Monte-Carlo estimate of the loss (\ref{eq:1}):
\begin{equation}\label{eq:2}
\hat{\bm{\Theta}}=\arg\min_{\bm{\Theta}}\sum_{i}l(\bm{y}_i-f(\bm{x}_i;\bm{\Theta}))\,.
\end{equation}
\par
This means that training an end-to-end network to approximate DSMs from their remote sensing images can result in estimating the expected height maps. But what is a good network architecture for our purpose?

\subsection{Architecture: Fully Convolutional-Deconvolutional Net}
\label{sec:netarchitecture}
The conventional CNNs are well known to be good at extracting features from data for concrete-to-abstract tasks like image classification~\cite{AlexNet,VGG,GoogLeNet}, by spatially shrinking the feature maps. In such networks, pooling is necessary to allow agglomerating information over the feature maps and, more fundamentally, to make the network computationally feasible. However, this produces feature maps with a reduced spatial resolution, so in order to provide dense per-pixel height maps we need to find a way to refine the coarse pooled feature representations.
\par
Fully convolutional network (FCN) has recently been actively studied for dense pixel-wise prediction tasks, e.g., semantic segmentation~\cite{FCN,Noh15}, image super-resolution~\cite{Dong16,Shi16}, and depth estimation~\cite{Eigen14,Eigen15}. To refine the downsampled output issue caused by the pooling operations in the conventional CNN framework, ``interpolation'' operations are usually used. For example, in~\cite{FCN}, Long \emph{et al.} present a method to iteratively refine the coarse feature maps, by applying the upsampling to the training phase instead of simply taking it as a post-processing step. This work exhibits that fully convolutional layers are capable of being replaced with convolutions whose filter size matches the layer input dimension. By doing so, the network is able to work on arbitrarily sized images while generating a desired full resolution output. Moreover, the CNN model proposed by Eigen \emph{et al.}~\cite{Eigen14} for depth prediction also suffers from this upsampling problem: the estimated depth maps are only $\frac{1}{4}$-resolution of their original input images and with some border areas lost. They refine those coarse depth maps by training an additional network which gets the coarse prediction and the input image as inputs.
\par
In this paper, we propose to use a network that is composed of two parts, i.e., convolutional sub-network and deconvolutional sub-network (see Fig.~\ref{fig:net}). The former corresponds to feature extractor that transforms the input remote sensing image to high-level multidimensional feature representation, whereas the latter plays the role of a height generator that produces height map from the feature extracted from the convolutional sub-network. Unlike~\cite{Eigen14,FCN}, we deconvolve the whole coarse feature maps, rather than only processing the coarse prediction, and this allows to transfer more high-level information to the fine prediction.
\par
So far we have confirmed the network architecture we adopted, and the next question is to instantiate a network, i.e., identifying details of the architecture (e.g., network super-parameters and loss function). It can be seen from Eq.~(\ref{eq:1}) that $f$ does matter, which involves such details. The following text will show how we build and gradually refine the a network for the height map estimation task.
\par

\begin{figure}[!t]
\centering
\begin{minipage}{\columnwidth}
\centering
\subfigure[]{
\label{fig:block_1}
\includegraphics[height=4.1cm]{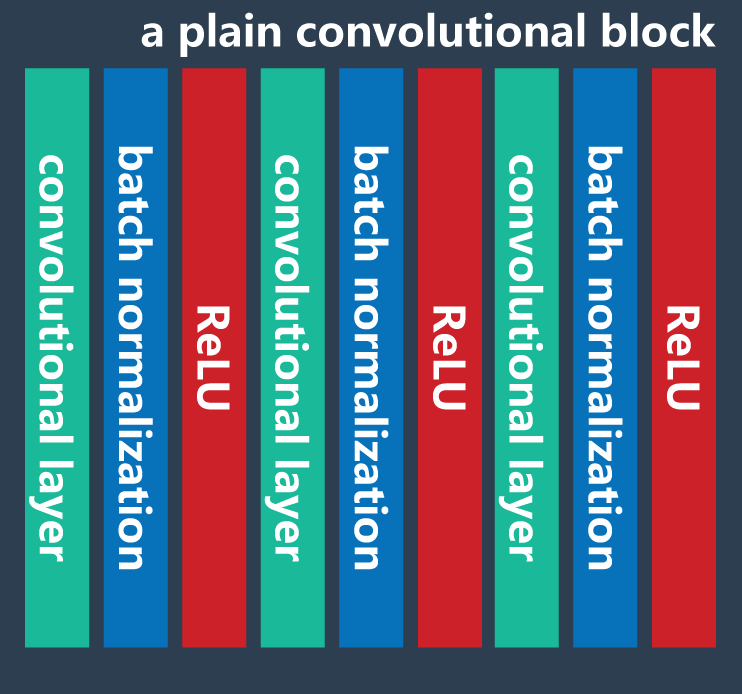}}
\subfigure[]{
\label{fig:block_2}
\includegraphics[height=4.1cm]{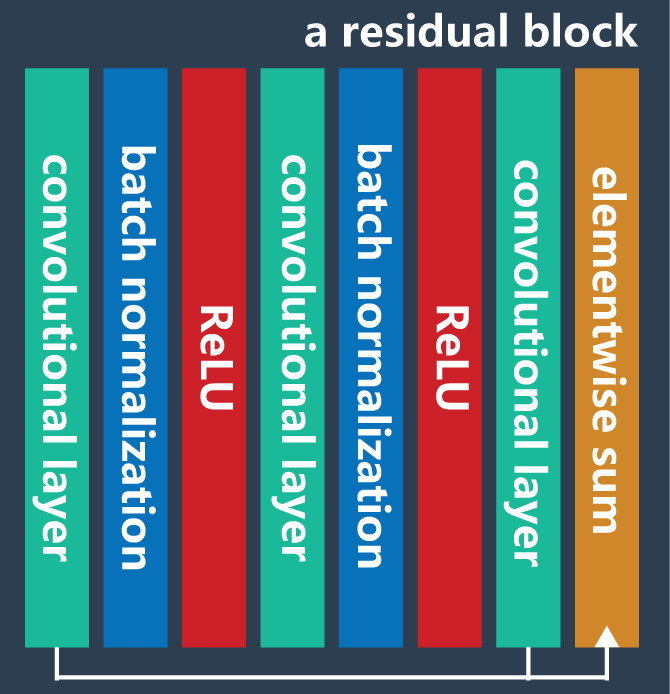}}
\end{minipage}
\renewcommand{\figurename}{Fig}
\caption{\label{fig:block} Illustration of plain convolutional block and residual block.}
\end{figure}

\subsection{Plain Convolutional Net vs. Residual Net}
\label{sec:convvsres}
The popular CNNs for dense per-pixel prediction tasks is VGG-like networks~\cite{FCN,Noh15,Johnson16,SegNet}. We, therefore, first attempt to build our fully convolutional-deconvolutional network using the philosophy of the VGG Nets~\cite{VGG}.
\par
The core component of the VGG Nets is plain convolutional block (cf. Fig.~\ref{fig:block_1}), which can make the networks simple and extensible. In general, our convolutional sub-network follows two rules of VGG Nets: 1) Having the same feature map size and the same number of filters in each convolutional layer of the same plain convolutional block; and 2) increasing the size of the feature maps in the deeper layers, roughly doubling after each max-pooling layer. The traits of the convolutional sub-network can be summarized as follows:
\begin{itemize}
  \item The input remote sensing image is fed into a stack of plain convolutional blocks, where we make use of convolutional filters with a very small receptive field of $3\times 3$, rather than leveraging larger ones, such as $7\times 7$ or $5\times 5$. That is because stacking $3\times 3$ convolutional layers can increase the nonlinearities inside the network.
  \item The convolutional stride is fixed to 1 pixel; the spatial padding is such that the spatial resolution of feature maps is preserved after convolution, i.e., the padding is 1 pixel for the used $3\times 3$ convolutional layers. Max-pooling is performed over $2\times 2$ pixel windows with stride 2.
\end{itemize}
\par

\begin{figure}[t]
\centering
\begin{minipage}{\columnwidth}
\centering
\subfigure[]{
\label{fig:unpooling_1}
\includegraphics[width=0.8\columnwidth]{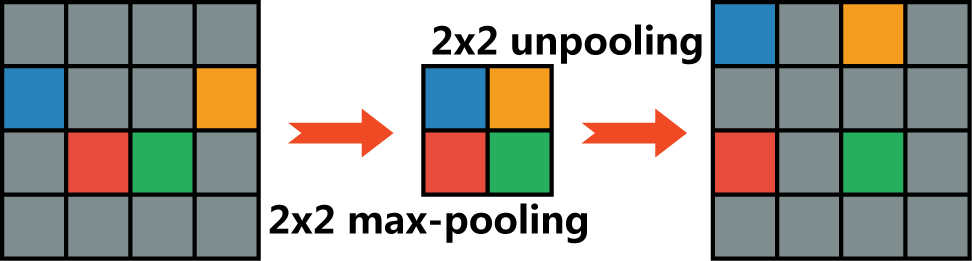}}
\subfigure[]{
\label{fig:unpooling_2}
\includegraphics[width=0.8\columnwidth]{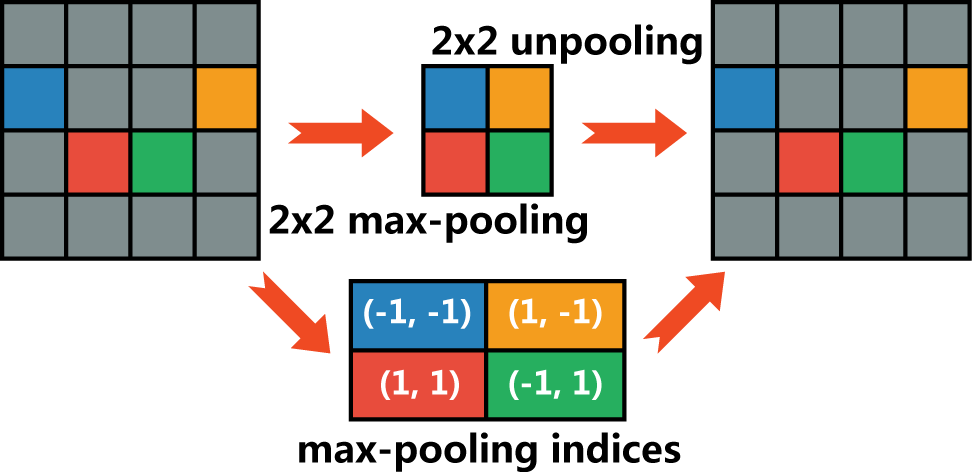}}
\end{minipage}
\renewcommand{\figurename}{Fig}
\caption{\label{fig:unpooling} An illustration of two types of unpooling operations: (a) simply replacing each entry of a feature map by an $s\times s$ block with the entry value in the top left corner and zeroing elsewhere; and (b) making use of max-pooling indices computed in the max-pooling layers of the convolutional sub-network.}
\end{figure}

\begin{figure*}[!t]
\begin{minipage}[t]{\linewidth}
\centering
\subfigure[]{
\includegraphics[width=0.187\linewidth]{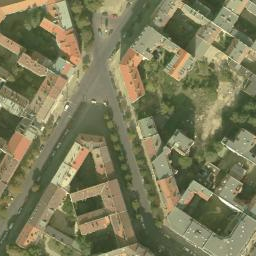}}
\subfigure[]{
\includegraphics[width=0.187\linewidth]{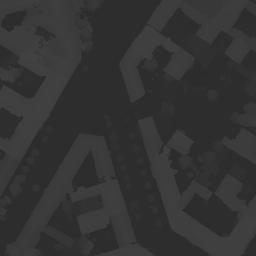}}
\subfigure[]{
\label{fig:plain_cd_example}
\includegraphics[width=0.187\linewidth]{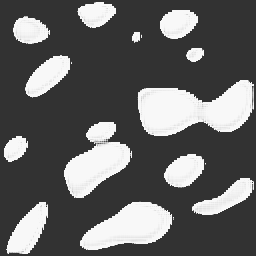}}
\subfigure[]{
\label{fig:res_cd_example}
\includegraphics[width=0.187\linewidth]{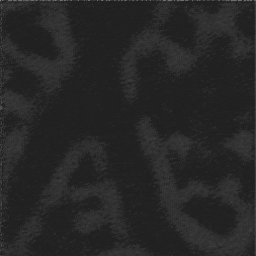}}
\subfigure[]{
\includegraphics[width=0.187\linewidth]{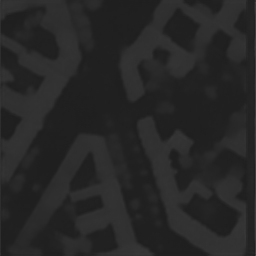}}
\end{minipage}
\renewcommand{\figurename}{Fig}
\caption{\label{fig:example_skip} Adding skip connection to the res. conv-deconv net to create a height map in much higher quality. \textbf{From left to right:} (a) remote sensing image, (b) ground truth, (c) plain conv-deconv net, (d) res. conv-deconv net, and (e) res. conv-deconv net with skip connection.}
\end{figure*}

The deconvolutional sub-network is a mirrored version of the convolutional sub-network, and its main ingredient is deconvolutional operation, which performs reverse operation of the convolutional sub-network and construct the height map from the abstract feature representation. The deconvolutional operation consists of \emph{unpooling} and \emph{convolution}. In order to map the encoded feature maps to a desired full resolution height map, we need \emph{unpooling} to unpool the feature maps, i.e., to increase their spatial span, as opposed to the pooling (spatially shrinking the feature maps). Dosovitskiy \emph{et al.}~\cite{Dosovitskiy15,Dosovitskiy16} perform \emph{unpooling} by simply replacing each entry of a feature map by an $s\times s$ block with the entry value in the top left corner and zeroing elsewhere (see Fig.~\ref{fig:unpooling_1}). In~\cite{Goroshin15,SegNet}, another form of \emph{unpooling} is implemented by making use of max-pooling indices computed in the max-pooling layers of the convolutional sub-network (cf. Fig.~\ref{fig:unpooling_2}). In this paper we choose the latter one as \emph{unpooling} strategy in our model, as the use of max-pooling indices theoretically enables location information to be more accurately represented and thus improves boundary delineation. Moreover, we achieve the \emph{convolution} using the same plain convolutional block as the convolutional sub-network.
\par
Now we have a reasonable network to handle our task, but a problem arises when we try to train it. The fully convolutional-deconvolutional network based on the plain convolutional blocks, which we will call \emph{plain conv-deconv net} hereafter, can reduce errors on both the training and validation sets during the first few iterations, but rapidly converges to a relatively high error value, which means it is not easy to optimize such a network. Furthermore, we also observe that the plain conv-deconv net fails to learn to produce physically plausible height maps (see an example in Fig.~\ref{fig:plain_cd_example}). To resolve this problem, we need to find a better way to construct the network.
\par

\begin{figure}[t]
\centering
\includegraphics[width=0.85\columnwidth]{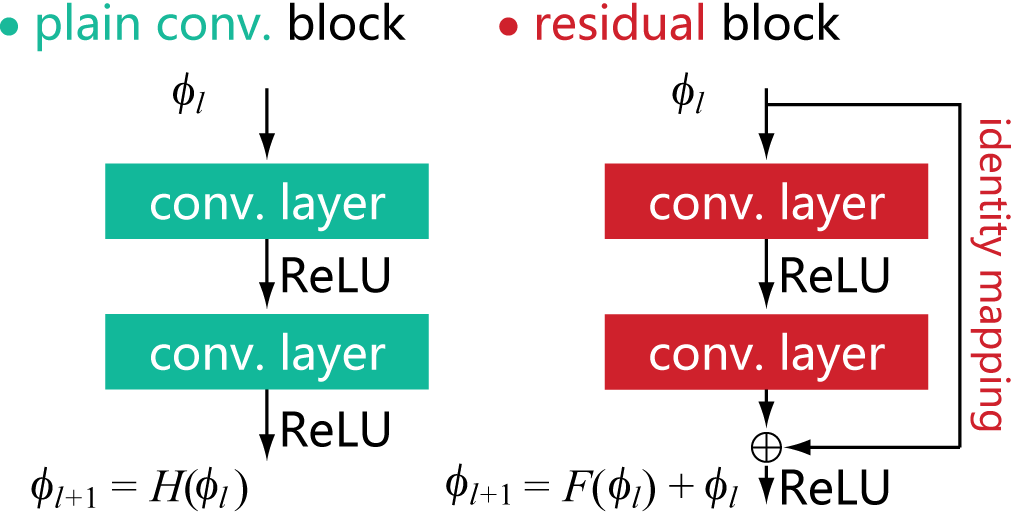}
\renewcommand{\figurename}{Fig}
\caption{\label{fig:convvsres} Comparison between the plain convolutional block and the residual block. Here $\bm{\phi}_l$ denotes the input and $\bm{\phi}_{l+1}$ is any desired output. The plain convolutional block hopes two convolutional layers are able to fit $\bm{\phi}_{l+1}$ by directly learning a mapping $\mathcal{H}$. In contrast, in the residual block the two convolutional layers are expected to learn a residual function $\mathcal{F}$ to let $\bm{\phi}_{l+1}=\mathcal{F}(\bm{\phi}_l)+\bm{\phi}_l$.}
\end{figure}

Recently, ResNet~\cite{ResNet} has achieved state-of-the-art results in image classification, winning ILSVRC 2015 with an incredible error rate of 3.6\%. The core idea of ResNet is building residual block (cf. Fig.~\ref{fig:block_2}), i.e., adding shortcut connections that by-pass two or more stacked convolutions by performing identity mapping and are then added together with the output of stacked convolutions. In~\cite{ResNet}, He \emph{et al.} show that ResNet is easier to optimize than plain networks like VGG Nets. In this paper, to solve the network training problem, we would like to introduce the residual learning to our convolutional-deconvolutional network architecture.
\par
Our fully residual convolutional-deconvolutional network (\emph{res. conv-deconv net} for short) is a modularized network that stacks residual blocks. Similarly to the plain convolutional block, a residual block consists of several convolutional layers that are with the same feature map size and have the same number of filters. However, it performs the following calculation:
\begin{equation}\label{eq:res1}
\bm{\varphi}_l=g(\bm{\phi}_l)+\mathcal{F}(\bm{\phi}_l;\bm{\Theta}_l)\,,
\end{equation}
\begin{equation}\label{eq:res2}
\bm{\phi}_{l+1}=\sigma(\bm{\varphi}_l)\,.
\end{equation}
\par
Here, $\bm{\phi}_l$ indicates the feature maps that are fed into the $l$-th residual block and satisfies $\bm{\phi}_0=\bm{x}$ where $\bm{x}$ is the input remote sensing image. $\bm{\Theta}_l=\{\bm{\Theta}_{l,k}|1\leq k\leq K\}$ represents a collection of weights associated with the $l$-th residual block, and $K$ denotes that there are $K$ convolutional layers in a residual block. Moreover, $\mathcal{F}$ is the residual function and is generally achieved by few stacked convolutional layers. The function $\sigma(\cdot)$ indicates the activation function such as a linear activation function or ReLU, and $\sigma(\cdot)$ works after element-wise addition. The function $g$ is fixed to an identity mapping: $g(\bm{\phi}_l)=\bm{\phi}_l$.
\par
If $\sigma(\cdot)$ adopts a linear activation function and also acts as an identity mapping, i.e., $\bm{\phi}_{l+1}=\bm{\varphi}_l$, we can obtain the output of the $l$-th residual block by putting Eq.~(\ref{eq:res1}) into Eq.~(\ref{eq:res2}):
\begin{equation}
\bm{\phi}_{l+1}=\bm{\phi}_l+\mathcal{F}(\bm{\phi}_l;\bm{\Theta}_l)\,.
\end{equation}
\par
In contrast, a plain convolutional block performs the following computation:
\begin{equation}
\bm{\phi}_{l+1}=\mathcal{H}(\bm{\phi}_l;\bm{\Theta}_l)\,.
\end{equation}
\par
Fig.~\ref{fig:convvsres} illustrates the difference between the plain convolutional block and the residual block, which is the latter is about learning residual instead of learning a complete mapping. After introducing the idea of residual learning to the network architecture, it can be clearly seen that the network becomes easier to be optimized (cf. Fig.~\ref{fig:res_cd_example}).

\begin{figure*}[!t]
\centering
\includegraphics[width=\linewidth]{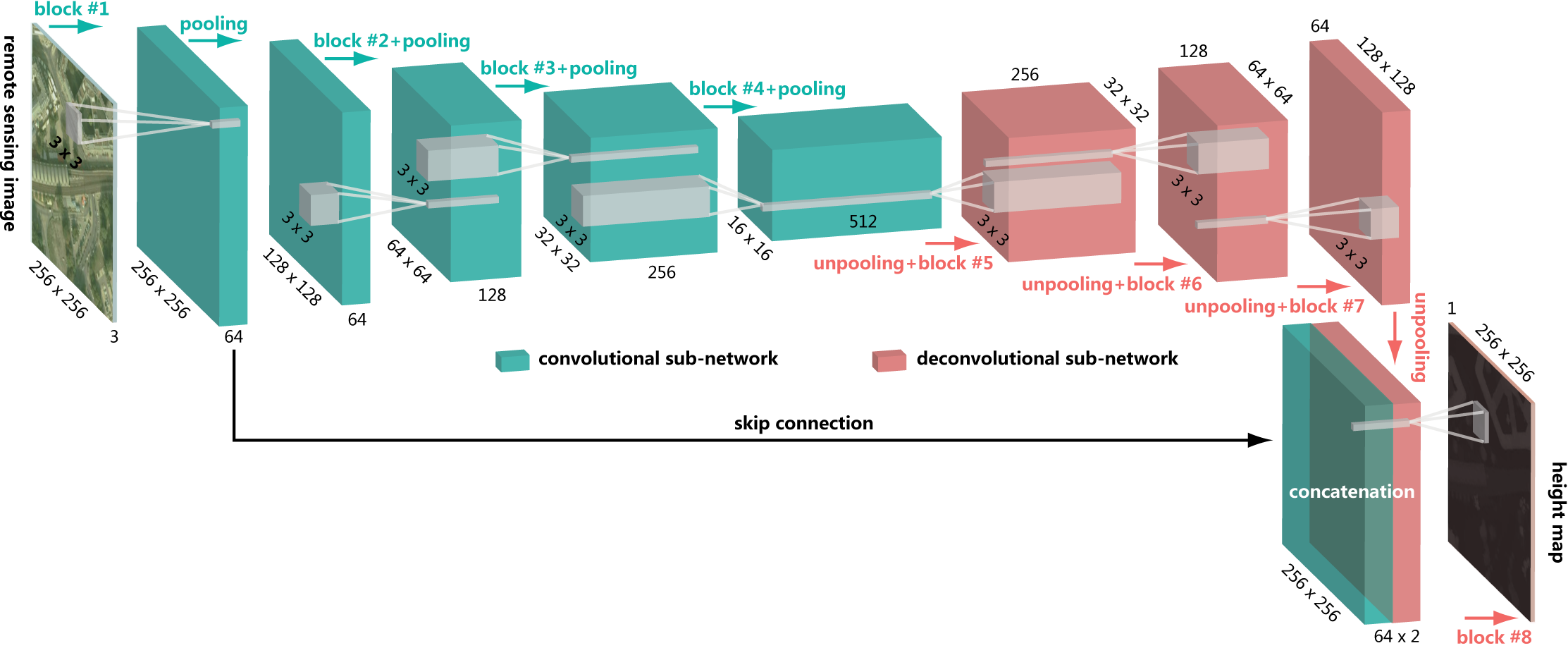}
\renewcommand{\figurename}{Fig}
\caption{\label{fig:net_skip} Overview of our final network architecture. We refine the network shown in Fig.~\ref{fig:net} by adding skip connection between the first block and the next to last block. This architecture makes it possible to shuttle low-level information, including object boundaries and edges, directly across the network. We adopt residual block (see Fig.~\ref{fig:block_2}) in this final network.}
\end{figure*}

\subsection{Res. Conv-Deconv Net with Skip}
\label{sec:skip}
For the problem we consider, the input high-resolution remote sensing image and output high-resolution height map differ in surface appearance, but both are renderings of the same underlying structure. Structure in the remote sensing image, therefore, should be roughly aligned with structure in the height map.
\par
In the res. conv-deconv net discussed above, the input image is passed through a stack of residual blocks that downsample inch by inch, until a bottleneck layer, at which point the process is reversed. Such a network architecture requires that all information flow pass through all the layers, including the bottleneck. For our problem, there is a great deal of low-level visual information, e.g., edges, shared between the input remote sensing image and output height map, and it would be desirable to shuttle such low-level information directly across the network. Unfortunately, due to the bottleneck layer, this can not be achieved in the current version of the network, which leads to the result that object boundaries tend to be blurred (see Fig.~\ref{fig:res_cd_example}). Both of~\cite{Hariharan14} and~\cite{MouDFC16} try to resolve this problem by combining the coarse or blurred semantic segmentation map with a superpixel segmentation of the input image to restore accurate object edges. However, such a strategy cannot be integrated into a network being trained end-to-end.
\par
Recently, several studies~\cite{Zeiler14,Mahendran15} that attempt to reveal what learned by CNNs show that deeper layers make use of filters to grasp global high-level information while shallower layers capture local low-level details. As reported in~\cite{Zeiler14}, the first layer of CNNs is always to detect object boundaries and edges, and such information is exactly what is missing in our network. To give the network a means to circumvent the bottleneck for the low-level visual information, we add skip connection between the first residual block and the next to last block to build a new network, and the skip connection simply concatenates all feature maps at the next to last block with those at the first block. With this strategy, the low-level information can be directly propagated without concerning any weight layers, which implies that the information containing object boundaries and edges will not vanish in the deconvolutional sub-network. Fig.~\ref{fig:example_skip} compares the network with skip connection (cf. Fig.~\ref{fig:net_skip}) against the network in Fig.~\ref{fig:net} in terms of height map quality.

\begin{figure*}[!t]
\centering
\includegraphics[width=\linewidth]{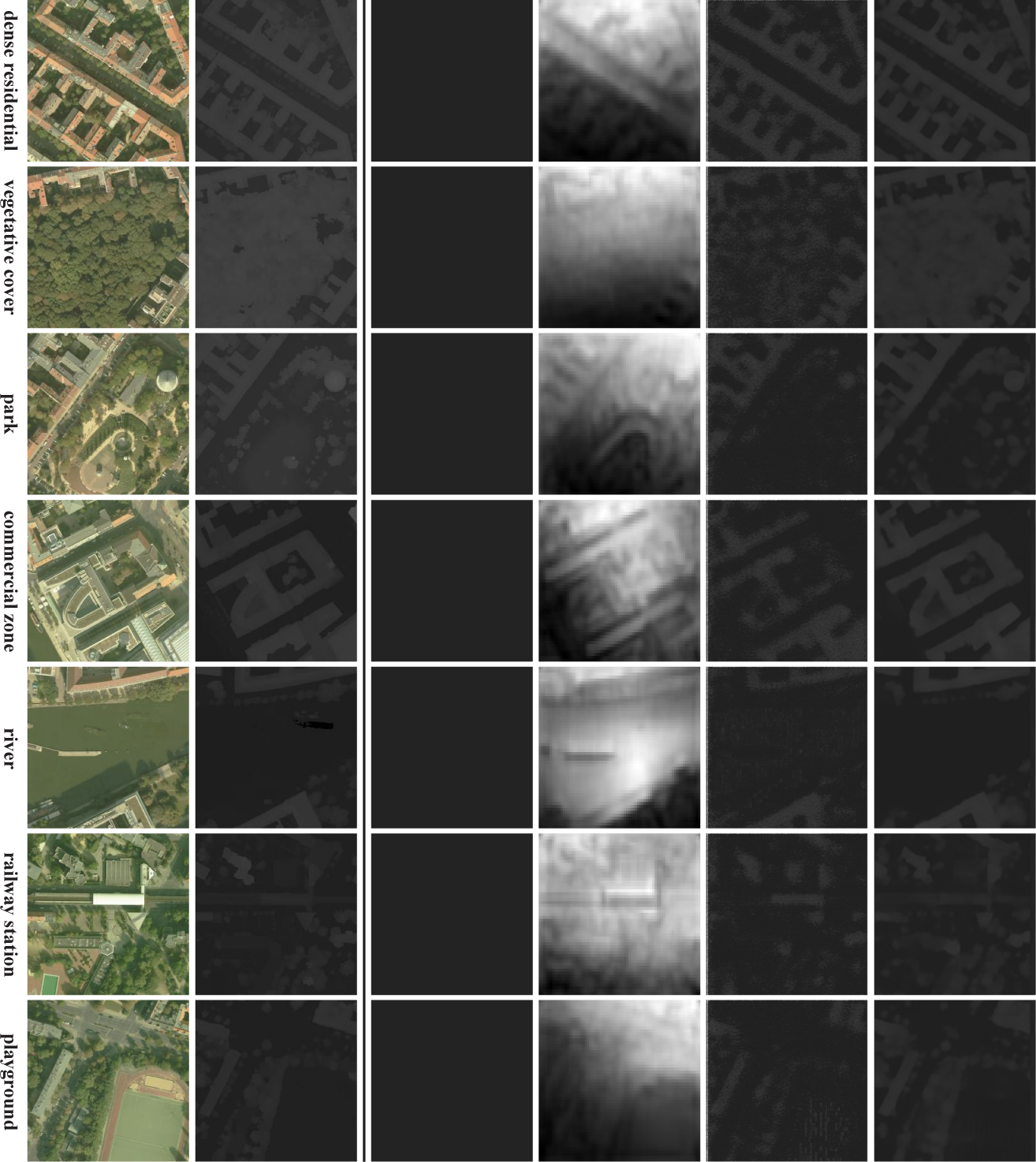}
\renewcommand{\figurename}{Fig}
\caption{\label{fig:zoom} Height prediction results on the different land use scenes of Berlin data set, considering $256\times 256$ pixels patches. \textbf{From left to right:} input image, ground truth, plain conv-deconv net with skip connection, Eigen-Net~\cite{Eigen14}, res. conv-deconv net, and res. conv-deconv net with skip connection.}
\end{figure*}

\section{Experiments}
\label{sec:exp}
\subsection{Data and Error Metrics}
\begin{figure*}[!t]
\begin{minipage}[t]{\linewidth}
\centering
\subfigure[]{
\label{fig:test_set}
\includegraphics[width=0.238\linewidth]{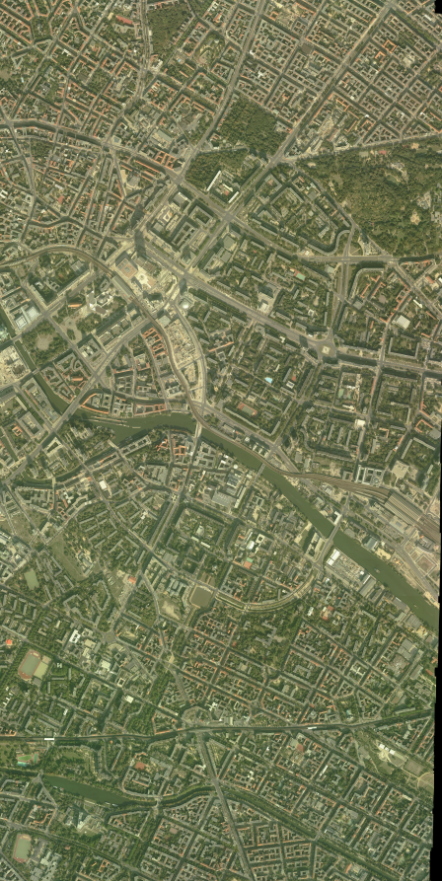}}
\subfigure[]{
\includegraphics[width=0.238\linewidth]{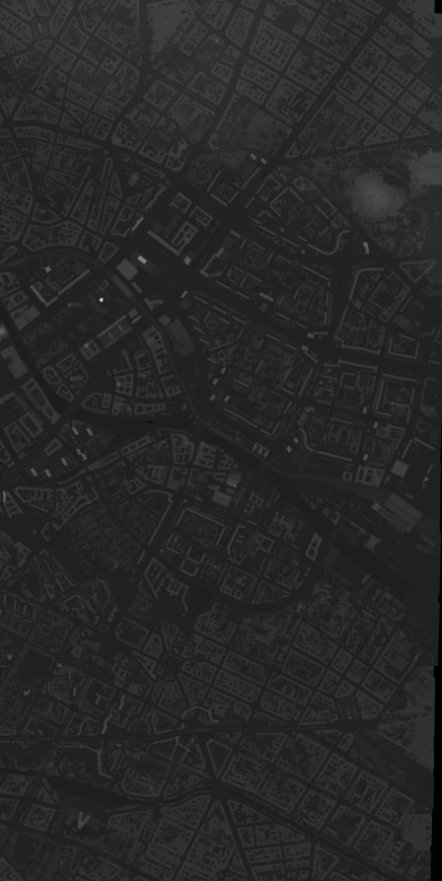}}
\subfigure[]{
\includegraphics[width=0.238\linewidth]{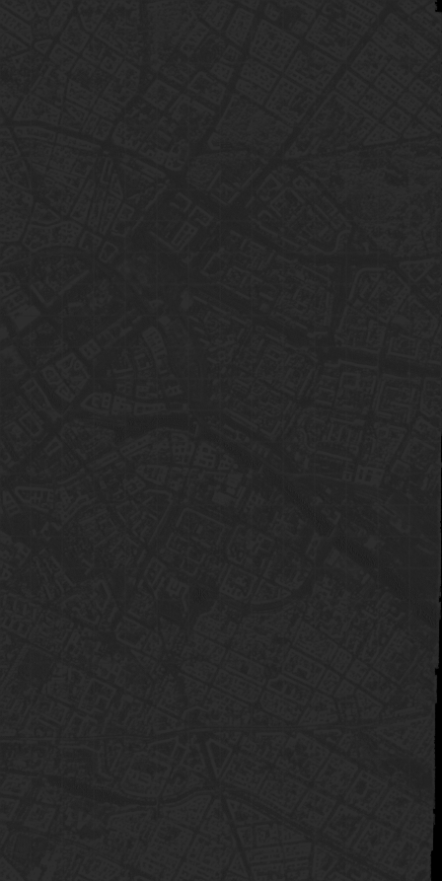}}
\subfigure[]{
\includegraphics[width=0.238\linewidth]{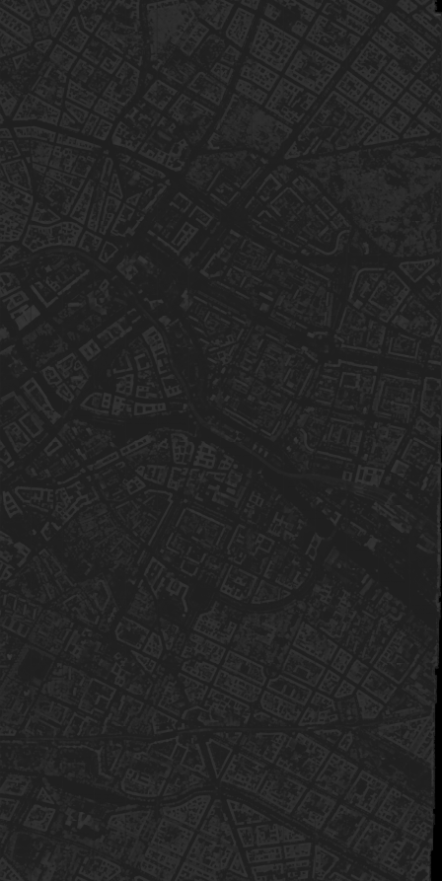}}
\end{minipage}
\renewcommand{\figurename}{Fig}
\caption{\label{fig:whole_scene} Height estimation on the large-scale Berlin scene ($3584\times 7168$). From left to right: (a) RGB image, (b) ground truth, (c) res. conv-deconv net, and (d) res. conv-deconv net with skip connection.}
\end{figure*}

We conduct our experiments on an aerial image over the area of Berlin, Germany. The DSM and the corresponding orthorectified aerial image were reconstructed using semi-global matching~\cite{Hirschmuller}. The data used in this research is resampled to about 70 cm in ground spacing. Results on this data are validated using $\left. 2 \middle / 3 \right.$ of the whole image for training and $\left. 1 \middle / 3 \right.$ for testing (see Fig.~\ref{fig:test_set} for the test area).
\par
To evaluate the performance of different approaches for height estimation from a single monocular image, a scale-invariant error metric is required. We adopt the scale-invariant error of~\cite{Eigen14}:
\begin{itemize}
  \item Mean squared error (MSE): $\frac{1}{n}\sum_j\|\bm{y}_j-\hat{\bm{y}}_j\|^2$.
  \item Mean absolute error (MAE): $\frac{1}{n}\sum_j|\bm{y}_j-\hat{\bm{y}}_j|$.
\end{itemize}

Moreover, in order to evaluate the estimated height map quality, another metric, structural similarity (SSIM) index, is also used in the experiment, which is given by~\cite{Wang04}
\begin{equation}
\frac{(2\mu_{\bm{y}}\mu_{\hat{\bm{y}}}+C_1)(2\sigma_{\bm{y}\hat{\bm{y}}}+C_2)}{(\mu_{\bm{y}}^2+\mu_{\hat{\bm{y}}}^2+C_1)(\sigma_{\bm{y}}^2+\sigma_{\hat{\bm{y}}}^2+C_2)}\,,
\end{equation}
where $\mu_{\bm{y}}$, $\mu_{\hat{\bm{y}}}$, $\sigma_{\bm{y}}$, $\sigma_{\hat{\bm{y}}}$, and $\sigma_{\bm{y}\hat{\bm{y}}}$ are the local means, standard deviations, and cross-covariance for images. SSIM is capable of comparing local patterns of pixel intensities that have been normalized for luminance and contrast.

\subsection{Training Details}
The network training is based on the Theano framework. We choose Nesterov Adam~\cite{nadam2,nadam1} as optimization algorithm to train our model from scratch, since for our task it shows much faster convergence than standard stochastic gradient descent with momentum~\cite{sgd} or Adam~\cite{adam}. We fix almost of all parameters of Nesterov Adam as recommended in~\cite{nadam2}: $\beta_1=0.9$, $\beta_2=0.999$, $\epsilon=1\mathrm{e}{-08}$, and a schedule decay of 0.004, but make use of a relatively small learning rate of 0.00002. All weights in the network are initialized with a Glorot normal initializer~\cite{Glorot_normal} that draws samples from a truncated normal distribution centered on zero. A standard loss function for optimization in regression problem is the $\ell_2$ loss, minimizing the squared Euclidean norm between predictions and ground truth. Although this loss performs well in our experiments, we find that using the $\ell_1$ loss yields a slightly better results. Hence, we use MAE as loss function in our network.
\par
The network is trained on RGB images to predict the corresponding height maps. The training set has only 840 unique $256\times256$ images. We use data augmentation to increase the number of training samples. The remote sensing images and corresponding DSM data are transformed using 1) rotation of input and target by 90 degree; and 2) horizontally and vertically flipping half of the images. By doing so, the number of training samples increase to 8,568. To monitor overfitting during training, we randomly select 10\% of the training samples as the validation set, i.e., splitting the training set into 7,711 training and 857 validation pairs. In addition, we use extremely small mini-batch of 1 image pair because, in a sense, every pixel is a training sample. Our network has 3,976,907 trainable parameters as well as 4,864 non-trainable parameters. Training a network of this size to give a good generalization error is very hard, therefore we make use of early stopping strategy to train it. We train our network on a single NVIDIA GeForce GTX TITAN with 12 GB of GPU memory and take about two hours.

\begin{table}[!t]
\caption{\label{tab:1} Statistics of Quantitative Estimation Performance on the Berlin Scene.}
\newcommand{\tabincell}[2]{\begin{tabular}{@{}#1@{}}#2\end{tabular}}
\centering
\begin{tabular}{cccc}
\toprule
Approach & MSE & MAE & SSIM \\
\hline
\hline
   res. conv-deconv net & 3.1e-03 & 2.7e-02 & 0.8060\\
   net with skip connection & 7.8e-04 & 1.7e-02 & 0.9366\\
\bottomrule
\end{tabular}
\end{table}

\subsection{Evaluation on Height Maps}
To demonstrate the effectiveness of the proposed network, we conduct experimental comparisons against several other networks. The approaches included in the comparison are listed as follows:
\begin{itemize}
  \item Eigen-Net: The network proposed in~\cite{Eigen14}. Note that the output height maps of this network are only $\frac{1}{4}$-resolution of their original input images. We, therefore, generate desired full resolution outputs using bilinear interpolation.
  \item Plain conv-deconv net with skip connection (see Fig.~\ref{fig:net_skip} and Fig.~\ref{fig:block_1}).%: The fully convolutional-deconvolutional network architecture with the plain convolutional blocks and the skip connections.
  \item Res. conv-deconv net (cf. Fig.~\ref{fig:net} and Fig.~\ref{fig:block_2}).%: A residual block-based fully convolutional-deconvolutional network.
  \item Res. conv-deconv net with skip connection. Fig.~\ref{fig:net_skip} and Fig.~\ref{fig:block_2} show the details.%: Our final network makes use of the residual blocks to construct the network and equips with the skip connection in order to shuttle low-level information directly across the network.
\end{itemize}
\par
In Fig.~\ref{fig:zoom} we qualitatively compare the accuracy of the estimated height maps using the proposed approach (res. conv-deconv net with skip connection) with that of the different variants (plain conv-deconv net with skip connection and res. conv-deconv net) as well as with the Eigen-Net on different land use scenes such as dense residential, vegetative cover, park, commercial zone, river, railway station, and playground. One can clearly see an improvement in quality from left to right. Compared to the Eigen-Net and res. conv-deconv net, the proposed model greatly improves the quality of object edges and boundaries and structure definition in the estimated height maps, which means that it can lead to better results, allowing low-level visual information to shortcut across the network.
\par
In addition, it can be seen from Fig.~\ref{fig:zoom} that the plain conv-deconv net with skip connection is unable to learn to produce realistic height maps in our experiments, and indeed collapses to generating nearly identical results for each input remote sensing image. Even if a couple of techniques such as Dropout~\cite{dropout} and $\ell_2$ regularization on network weights are adopted during training phase, we still cannot learn a valid network model. This proves that when both res. conv-deconv net and plain conv-deconv net are trained in our task, the former is capable of achieving the superior results, whereas the latter is pretty difficult to optimize. Moreover, Fig.~\ref{fig:zoom} shows that to a certain extent, Eigen-Net~\cite{Eigen14} is able to learn some structures in the estimated height maps, but these results are quite blurry. In contrast, our height predictions exhibit noteworthy visual quality, even though they are derived by a single network being trained end-to-end, without any additional post-processing steps.
\par

\begin{figure*}[!t]
\begin{minipage}[t]{\linewidth}
\centering
\subfigure[]{
\includegraphics[width=\linewidth]{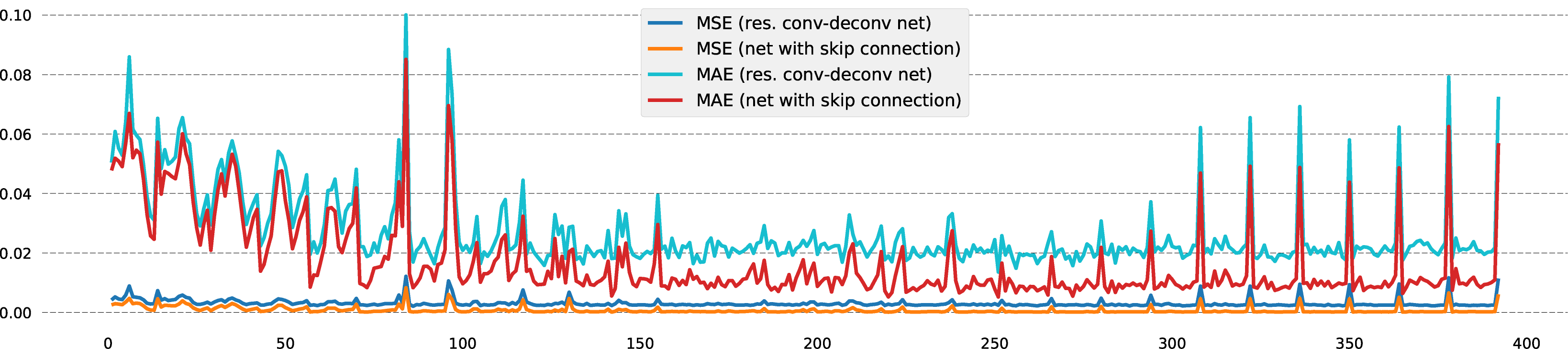}}
\end{minipage}
\begin{minipage}[t]{\linewidth}
\centering
\subfigure[]{
\includegraphics[width=\linewidth]{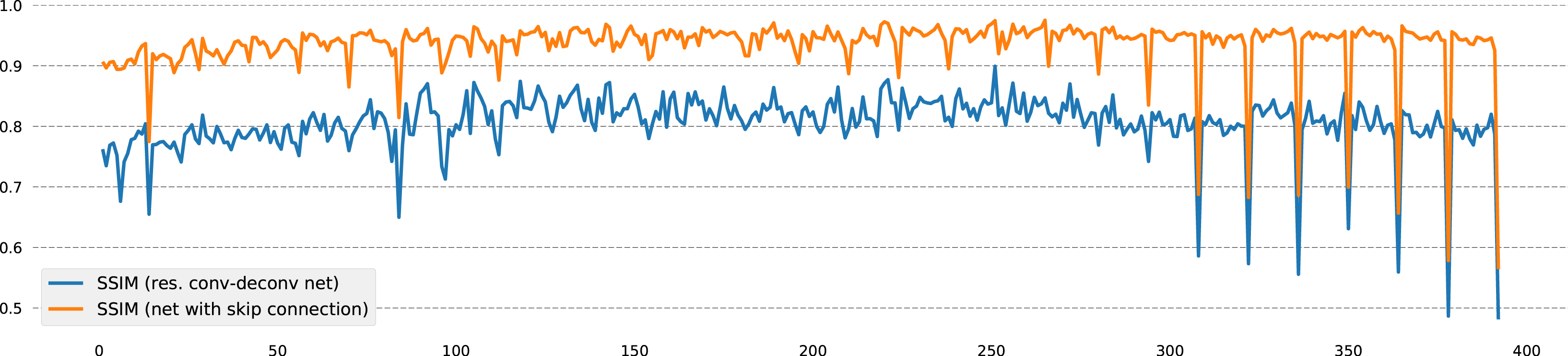}}
\end{minipage}
\renewcommand{\figurename}{Fig}
\caption{\label{fig:metrics} MSE, MAE, and SSIM of the res. conv-deconv net and
the net with skip connection on 392 unique $256\times 256$ local patches of the test scene. We randomly select several failure images (with fairly high error and low SSIM) and show them in Fig.~\ref{fig:failure}.}
\end{figure*}

\begin{figure}[t]
\centering
\includegraphics[width=\columnwidth]{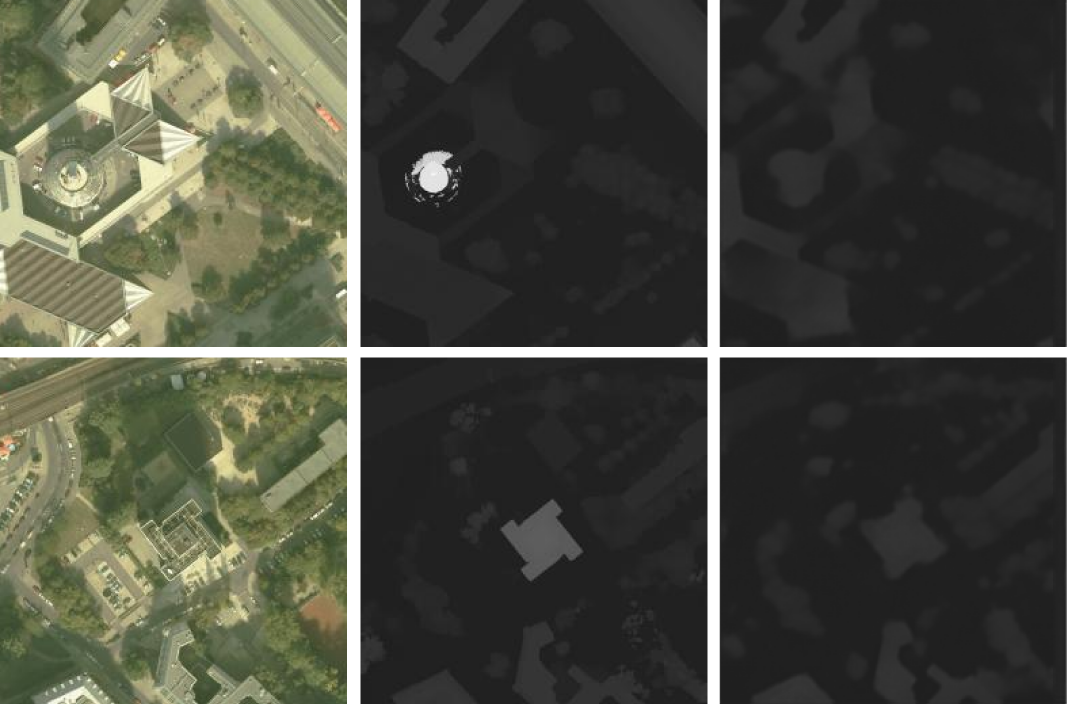}
\renewcommand{\figurename}{Fig}
\caption{\label{fig:failure} Example failure cases. \textbf{Left:} input remote sensing image. \textbf{Middle:} ground truth. \textbf{Right:} predicted height map. Common failures include difficulty in handing relatively high-rise buildings, that is the main challenge of predicting height from monocular orthographic remote sensing image.}
\end{figure}

\begin{figure}[t]
\centering
\includegraphics[width=\columnwidth]{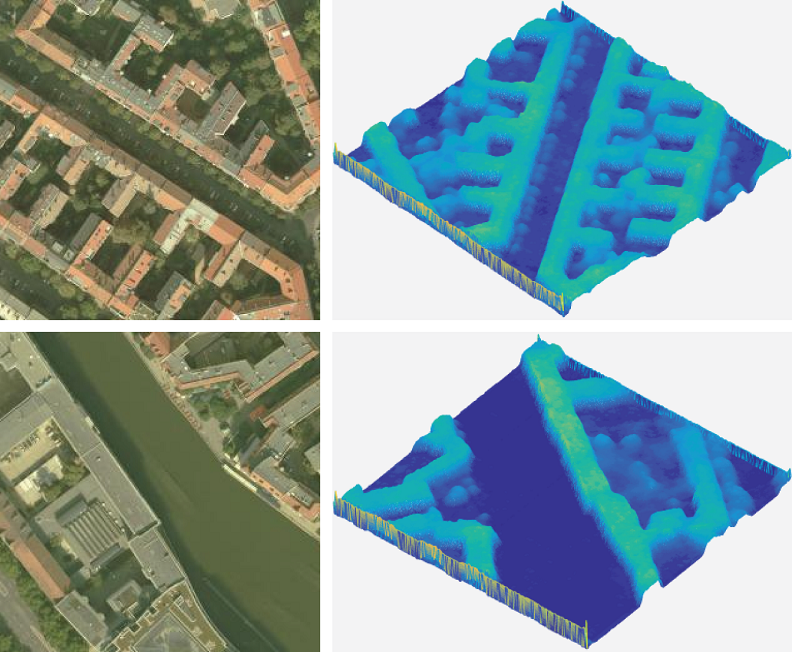}
\renewcommand{\figurename}{Fig}
\caption{\label{fig:3d} Point cloud visualization of the height estimates for two selected examples.}
\end{figure}

We do not have any fully connected layers in our network, which allows the network to take remote sensing images of arbitrary size as input. Fig.~\ref{fig:whole_scene} shows height estimation results on the large-scale Berlin test zone. As shown in this figure, the proposed network with skip connection can obtain a height map with very good low-level visual details (e.g., object boundaries) instead of a blurry one as do res. conv-deconv net. Table~\ref{tab:1} list the quantitative statistics on the test data set. Specifically, the res. conv-deconv net with skip connection increases the accuracy significantly by 0.0023 of MSE, 0.01 of MAE, and 0.1306 of SSIM, respectively. Furthermore, to better understand the performance of the networks, we divide the whole test image into 392 unique $256\times 256$ patches, and Fig.~\ref{fig:metrics} shows the performance of the res. conv-deconv net and the net with skip connection on these 392 scenes. We can see that introducing skip connection to the res. conv-deconv net strikingly helps with a large majority of scenes. However, the proposed network still does poorly on some complex scenes such as tall buildings. Thus, even if we could get good object boundaries and edges, the task of estimating many of the tall buildings would remain quite challenging. According to Fig.~\ref{fig:metrics}, we select several failure cases, as shown in Fig.~\ref{fig:failure}. In addition, the output of our model can be used to generate novel 3D views of the scene from a single monocular image (see Fig.~\ref{fig:3d}).

\subsection{Application to Instance Segmentation of Buildings}
To complement the previous results, in this section we show a practical case, instance segmentation of buildings, to demonstrate the usefulness of height estimation from a single monocular image. The goal of instance segmentation is to identify individual instances of buildings in pixel-level in an image. In general, most instance segmentation approaches rely on strong supervision for training, i.e., pixel-level segmentation masks. However, in comparison with convenient image-level labels, collecting annotations of pixel-wise segmentation ground-truth is much more expensive. In particular, for the instance segmentation task, labeling a considerable number of pixel-level segmentation masks usually requires a large amount of human effort as well as financial expenses.
\par
\begin{figure}[t]
\centering
\includegraphics[width=0.9\columnwidth]{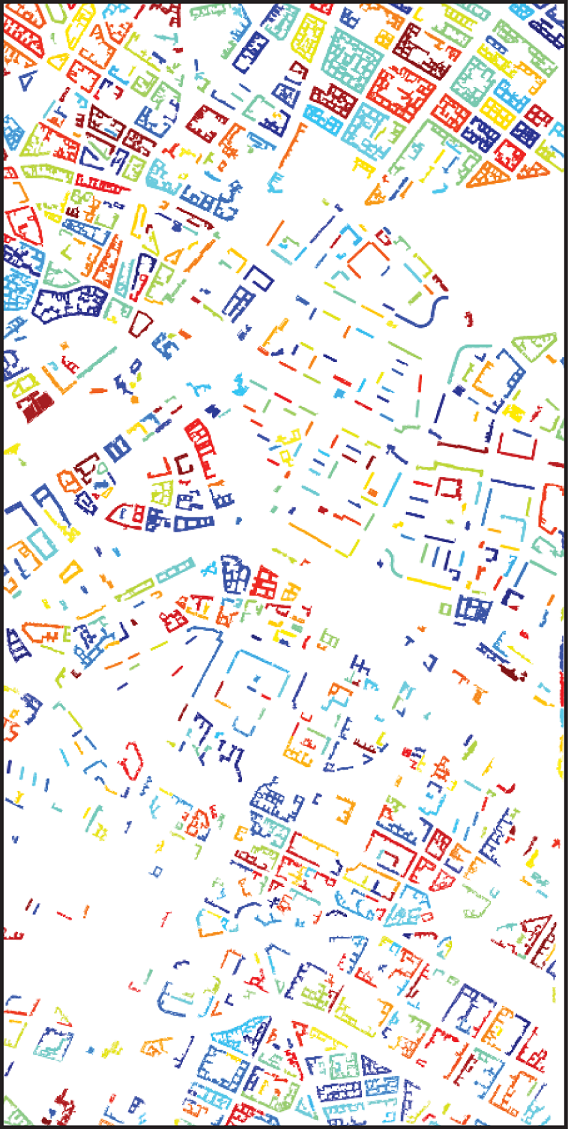}
\renewcommand{\figurename}{Fig}
\caption{\label{fig:ins_seg} The result of instance segmentation of buildings based on the predicted height map on the Berlin test scene. Two zoomed segmentation examples are shown in Fig.~\ref{fig:ins_seg_zoom}.}
\end{figure}
\begin{figure*}[t]
\centering
\includegraphics[width=0.9\linewidth]{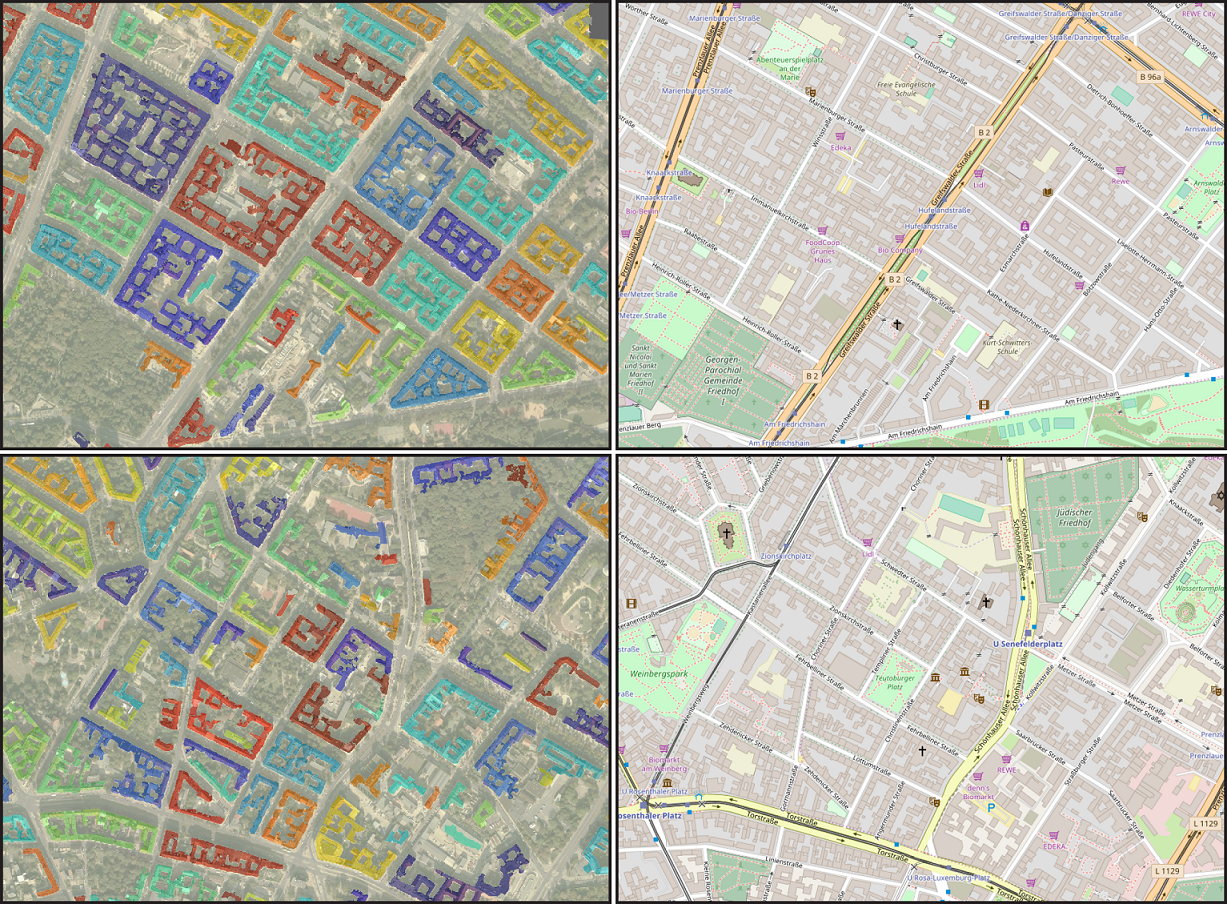}
\renewcommand{\figurename}{Fig}
\caption{\label{fig:ins_seg_zoom} \textbf{Left:} two close zooms of the instance segmentation map. \textbf{Right:} corresponding areas in open street map as reference.}
\end{figure*}

In this practical case, we give a different perspective to achieve instance segmentation of buildings, i.e., utilizing the estimated height map generated by the proposed res. conv-deconv net with skip connection. Specifically, we deploy an instance segmentation framework where structures elevated above the ground level (i.e., buildings and trees) are first extracted by setting a threshold in the predicted height map, and then trees are filtered out using vegetation index (VI). Finally, post-processing steps including removing small areas and filling holes are performed. We wish to point out that, to the best of our knowledge, this is the first demonstration of an instance segmentation of buildings based on height estimation from a single monocular image. The result on the Berlin test scene is shown in Fig.~\ref{fig:ins_seg}, and we also provides two close zooms of the instance segmentation map (cf.~\ref{fig:ins_seg_zoom}). As it can be seen, the segmentation result is satisfactory, especially considering the fact that our approach dose nor rely on any pixel-wise labeled mask and supervised training.

\section{Conclusion}
\label{sec:con}
In this paper we propose a fully residual convolutional-deconvolutional network in order to deal with a novel problem in the remote sensing community, i.e., single-view height prediction. In particular, the proposed network consists of two parts, namely convolutional sub-network and deconvolutional sub-network. They are responsible for transforming an input high resolution remote sensing image to abstract multidimensional feature representations and generating height map, respectively. However, during the experiment, we found that due to the bottleneck of the network, such a network easily leads to the result that object boundaries tend to be blurred. To address this problem, we refine the network architecture by adding a skip connection between the first residual block and the next to last block, which makes it possible to shuttle low-level information, e.g., object boundaries, directly across the network. in addition, in the experimental section we show an application to instance segmentation of buildings to demonstrate the usefulness of the predicted height map from a single monocular image. In the future, further experiments and studies will be focused on how to improve the accuracy of high-rise buildings.

\section*{Acknowledgement}
We thank H. Hirschm\"uller of Institute of Robotics and Mechatronics of DLR for providing the data used in this research.

\ifCLASSOPTIONcaptionsoff
  \newpage
\fi

\bibliographystyle{IEEEtran}
\bibliography{IEEEfull,IM2HEIGHT_v1}

\end{document}